\documentclass[letterpaper, 10 pt, conference]{ieeeconf}  

\IEEEoverridecommandlockouts
\usepackage{cite}
\usepackage{amsmath,amssymb,amsfonts}
\usepackage[ruled,longend]{algorithm2e}
\usepackage{graphicx}
\usepackage{textcomp}
\usepackage{xcolor}
\usepackage{eqexpl}
\usepackage{autobreak}
\usepackage{comment}

\usepackage{stfloats} 
\usepackage{amsmath}

\usepackage[utf8]{inputenc}

\usepackage{textgreek}
\usepackage{amssymb}
\usepackage{float}
\usepackage{svg}
\usepackage{tikz}
\usepackage{pgfplots}
\usepackage{booktabs}
\usepackage{authblk}
\usepackage{float}
\usepackage{mathtools}

\usepackage{mdframed} 

\usepackage[most]{tcolorbox}

\usepackage{algpseudocode}
\usepackage{caption}
\usepackage{url}

\graphicspath{{Figures/}}
\pgfplotsset{compat=1.17}

\newlength\imagewidth
\newlength\imagescale

\newcommand{\ra}[1]{\renewcommand{\arraystretch}{#1}}

\def\BibTeX{{\rm B\kern-.05em{\sc i\kern-.025em b}\kern-.08em
    T\kern-.1667em\lower.7ex\hbox{E}\kern-.125emX}}

\begin{document}

\title{HighwayLLM: Decision-Making and Navigation in Highway Driving with RL-Informed Language Model 

}

\author{Mustafa Yildirim$^{1}$, Barkin Dagda $^{1}$,  Saber Fallah$^{1}$
\thanks{$^{1}$Mustafa Yildirim, Barkin Dagda and Saber Fallah are with CAV-Lab, Department of Mechanical Engineering Sciences, University of Surrey,
        {\tt\small \{m.yildirim, bd00242, s.fallah\}@surrey.ac.uk}}%

}

\maketitle

\begin{abstract}
Autonomous driving is a complex task which requires advanced decision making and control algorithms. Understanding the rationale behind the autonomous vehicles' decision is crucial to ensure their safe and effective operation on highway driving.
This study presents a novel approach, \textit {HighwayLLM}, which harnesses the reasoning capabilities of large language models (LLMs) to predict the future waypoints for ego-vehicle's navigation. Our approach also utilizes a pre-trained Reinforcement Learning (RL) model to serve as a high-level planner, making decisions on appropriate meta-level actions. The HighwayLLM combines the output from the RL model and the current state information to make safe, collision-free, and explainable predictions for the next states, thereby constructing a trajectory for the ego-vehicle. Subsequently, a PID-based controller guides the vehicle to the waypoints predicted by the LLM agent. This integration of LLM with RL and PID enhances the decision-making process and provides interpretability for highway autonomous driving.
\end{abstract}


\section{Introduction}

Autonomous Driving (AD) indicates a significant shift in transportation, promising to improve efficiency, reduce accidents, and increase mobility. However, this arising technology faces major challenges, particularly in navigation. Autonomous vehicles (AVs) must accurately perceive their environment, make complex decisions in real time, and adapt to dynamic conditions while ensuring safety and compliance with traffic rules. A critical aspect of AD is its decision-making capability for lane changing and overtaking. Lane changing involves continuous observation of surrounding vehicles, assessing sight distance and velocity, and predicting intention of other vehicles to change lanes. Effective lane change can impact the efficiency of the vehicle's acceleration, velocity, time spent in traffic and safety. Inaccurate decision-making for a lane change can lead to traffic congestion, increased risk of accidents, and reduced passenger comfort.


The control for automated lane change has been studied in various approaches. Rule-based methods such as Mobil \cite{kesting2007general} explicitly formulate a lane change based on the gap and velocity of vehicles at the desired lane. Machine learning methods such as random forest \cite{arbabi2020lane} rely on static models that cannot fully capture the dynamic complexities and unpredictable behaviours in real-world driving scenarios. Deep Reinforcement Learning has been adapted to learn a safe and efficient lane change by high-level lateral decision-making \cite{alizadeh2019automated,yildirim2022prediction} with a low-level rule-based lane change process. For example, Shi et al. \cite{shi2019driving} used Deep Q-network (DQN) to learn when to do lane change manoeuvres. However, existing RL-based approaches are black boxes that do not reason their performed actions \cite{wells2021explainable}. 
In such driving conditions, understanding the rationale behind an AV's decisions is essential for maintaining safety and efficiency on the road. Based on a survey conducted by Othman et al. \cite{othman2021public}, 74\% of respondents do not trust AVs or believe that AVs are not better than human drivers. The survey also reports that 57\% of the participants have concerns about AV awareness.

Explainable RL \cite{wells2021explainable} provides reasoning behind actions to humans, making the decision-making process transparent and understandable for humans. This transparency is crucial for passengers in making understandable decisions and providing insights into the decision-making process of AVs \cite{atakishiyev2021explainable}. Especially in critical scenarios, understanding the rationale behind an AV's decisions is essential for ensuring safety and accountability.
Reasoning and understanding actions's cause is a necessity for a human nature but for autonomous driving this is not at the desired level \cite{nastjuk2020drives}.
In some applications, a vehicle onboard screen shows planned route but does not provide a rationale about its action and explanation for future actions \cite{LSE2019TrustInAutomation}. Therefore a better explainability and reasoning is required for gaining passengers' trust.

RL has shown promise in addressing the complexities of autonomous vehicle navigation. It enables an agent to learn optimal behaviour through a trial-and-error approach by interacting with its environment. RL can be utilized to make decisions, adapt to changing traffic scenarios and learn from past experiences. 
Proportional-Integral-Derivative (PID) is an effective controller to navigate a vehicle to change lane in a given set of vehicle parameters and target in dynamic environments, making it suitable for path tracking and control.

LLMs \cite{arisoy2012deep,vaswani2017attention,devlin2018bert} are designed to understand and interpret natural language as a human does. This paper introduces a novel language model-based trajectory planner for autonomous highway driving. Our approach integrates RL and PID with an LLM agent to predict the future states of the ego-vehicle with reasoning provided for its actions. Our proposed approach bridges the gap between human understanding and black-box RL decision-making for autonomous driving.

The integration of RL for decision-making and PID for vehicle navigation, informed by predictive trajectories from an LLM-based API, presents a novel approach that could address the existing gaps in autonomous vehicle navigation systems.
Contribution of this study:
\begin{itemize}
    \item To the best of our knowledge, for the first time, we use an LLM as a trajectory planner for a highway driving scenario, and we reason the decisions of the RL agent for a lane change in a natural language.
    \item We build a multi-modal trajectory planner by combining RL, PID and LLM. 
    \item We integrated RL and LLM and tested interactively in real-time simulation based on real traffic dataset.
\end{itemize}

\section{Related Work}

\subsection{Decision Making for Autonomous Driving}

Decision-making in AD involves observing surrounding vehicles and understanding and predicting their future actions. Different methods have been implemented in the literature, but each method has a drawback.
Rule-based methods are not generalisable, and one can not generate rules for every instance. Due to complexity and computational expense, Fuzzy Logic \cite{naranjo2008lane}, or Monte Carlo Tree Search \cite{hoel2019combining}, are not scalable to multi-vehicle environments. Supervised learning methods \cite{kuutti2022end} require large data to train. RL methods\cite{alizadeh2019automated,arbabi2020lane} show promising results in autonomous driving and outperform the other methods, but they require many trial-error computations. A further disadvantage of RL methods is that they are, by nature, black-box methods \cite{hassija2023interpreting} and cannot be explained.


\subsection{Large Language Models}
Large language models (LLMs) are large artificial intelligence (AI) models, designed to understand, generate, and interpret human language. The recent LLMs such as ChatGPT \cite{radford2018improving}, the open-source Llama \cite{touvron2023llama} and Mistral of Experts \cite{jiang2023mistral}, have demonstrated superior capabilities in various natural language processing (NLP) tasks.

The application of LLMs also expanded into the robotics and autonomous driving. Recent studies, such as Mao et al. \cite{mao2023gpt} fine-tuned GPT 3.5 to act as a high level motion planner by generating trajectories. The authors evaluate their approach on UniAD dataset \cite{hu2023_uniad} with few-shot learning performance of the GPT. In another study, Xu et al. \cite{xu2023drivegpt4} propose DriveGPT4, an end-to-end pipeline for predicting low level control signals by interpreting the video frames and corresponding human generated questions. The authors evaluate the zero-shot performance of their vision language model approach and report significant generalisation for daily driving applications. DriveGPT4 exemplifies how LLMs can be integrated into the operational frameworks of autonomous driving systems, enhancing their ability to provide understandable and interpretable decision-making processes.

In this paper, we propose an RL-informed LLM agent that acts as a motion planner in a highway driving. The highway driving is a complex and unpredictable environment. Our proposed \textit{HighwayLLM} uses the pretrained RL model to predict the meta level actions and our LLM agent uses the output from the RL along with the current state information and past similar trajectories retrieved from a large database to make safe and collision free predictions for the ego-vehicle.

\section{Proposed Approach}
\subsection{Reinforcement Learning Model}
In this study as a decision making model we have implemented Deep Q Networks (DQN)\cite{mnih2013playing}. Markov Decision Processes (MDPs) are mathematical frameworks for decision-making that are used in the formulation of RL problems\cite{Bellman1977a}. It consists of a set of actions $\mathcal{A}$, a set of states $\mathcal{S}$, a reward function $\mathcal{R}$, and a transition model $P(s'|s, a)$. In an MDP context, an RL agent aims to optimise its policy $\pi$ to maximise the sum of expected rewards over time. This is achieved by selecting an action $a \in \mathcal{A}$, transitioning to a new state $s' \in \mathcal{S}$, and obtaining a reward $r$. To maximise future cumulative rewards, the agent updates its policy $\pi$ until it has sufficiently optimised the policy to achieve the highest possible score.

Cumulative rewards represent the total rewards received over a series of events, with a discount factor $\gamma \in [0,1]$ applied at each step to trade off the significance of immediate versus future rewards.

The state space includes the position and velocity of the ego vehicle (EV), as well as the distance and speed of the surrounding vehicles. EV's action space consists of left lane changes, right lane changes, or lane-keeping A=\{LLC, RLC, LK\}. RL aims to determine the optimal policy $\pi$ that maximizes the sum of all future rewards:

\begin{equation}
R(\pi, r)=\mathbb{E}_{\pi}\left[\sum_{t=0} \gamma^{t} r\left(s_{t}, a_{t}\right)\right]
\end{equation}

In solving sequential decision problems, a crucial step is learning estimations for the optimal value of each action. This is defined as the expected value of the rewards in the future when taking that action and following the optimal policy accordingly. The policy predicts the value for each action in A for the current state\cite{yildirim2022prediction}. This significant concept, known as the action-value function, is formulated as:

\begin{equation}
    Q_{\pi}(s,a)=\mathbb{E}[R_{1} + \gamma R_{2} + ... | S_0 = s, A_0 = a, \pi]
\end{equation}


Using this action-value function, actions can be chosen greedily or exploratory based on the current observation s $\epsilon$ S.

After each iteration ${i}$, the loss function below is minimized by optimizing the network weights $\theta$,

\begin{equation}
\begin{aligned}
L_{i}\left(\theta_{i}\right)=\mathbb{E}_{s, a \sim \rho(\cdot)}\left[\left(y_{i}-Q\left(s, a ; \theta_{i}\right)\right)^{2}\right]\\
y_{i}=\mathbb{E}_{s^{\prime} \sim \varepsilon}\left[r+\gamma \max _{a^{\prime}} Q\left(s^{\prime}, a^{\prime} ; \theta_{i-1}\right) \mid s, a\right]
\label{eqn:Function Aproximation Loss}
\end{aligned}
\end{equation}

$y_i$ is the target and $\rho(s,a)$ is the probability distribution of states and actions. The differential of the loss function with respect to the weights provides the following gradient.

\begin{multline}
\nabla_{\theta_{i}} L_{i}\left(\theta_{i}\right)=\mathbb{E}_{s, a \sim p(\cdot) ; s^{\prime} \sim \mathcal{E}} 
\bigg[\left(r+\gamma \max _{a^{\prime}} Q\left(s^{\prime}, a^{\prime} ; \theta_{i-1}\right)\right) \\-
Q\left(s, a ; \theta_{i}\right) \nabla_{\theta_{i}} Q\left(s, a ; \theta_{i}\right)\bigg]
\label{eqn:Loss Gradient}
\end{multline}

This gradient is followed through Adam (Adaptive Moment Estimation) gradient decent on each iteration.

\subsection{Vehicle Model}

In this study we have modeled vehicle as a unicycle model. The state of the vehicle represented by both its lateral and longitudinal positions and velocities. The control inputs are the lateral and longitudinal accelerations. The  vehicle as a unicycle model as defined in equations \ref{eq:motion_equations} are:

\begin{equation}
\begin{aligned}
y_{\text{t+1}} &= y + \Delta t \cdot v_{y}, \\
v_{y_\text{t+1}} &= v_{y} + \Delta t \cdot a_{y}, \\
x_{\text{t+1}} &= x + \Delta t \cdot v_{x}, \\
v_{x_\text{t+1}} &= v_{x} + \Delta t \cdot a_{x}.
\end{aligned}
\label{eq:motion_equations}
\end{equation}

Where \( y \) is the lateral position, \( v_{y} \) is the lateral velocity, \( a_{y} \) is the lateral acceleration, \( x \) is the longitudinal position, \( v_{x} \) is the longitudinal velocity, \( a_{x} \) is the longitudinal acceleration and \( \Delta t \) is the time step.



\begin{figure*}
\centering
\includegraphics[width=1\textwidth,height=\textheight,keepaspectratio]{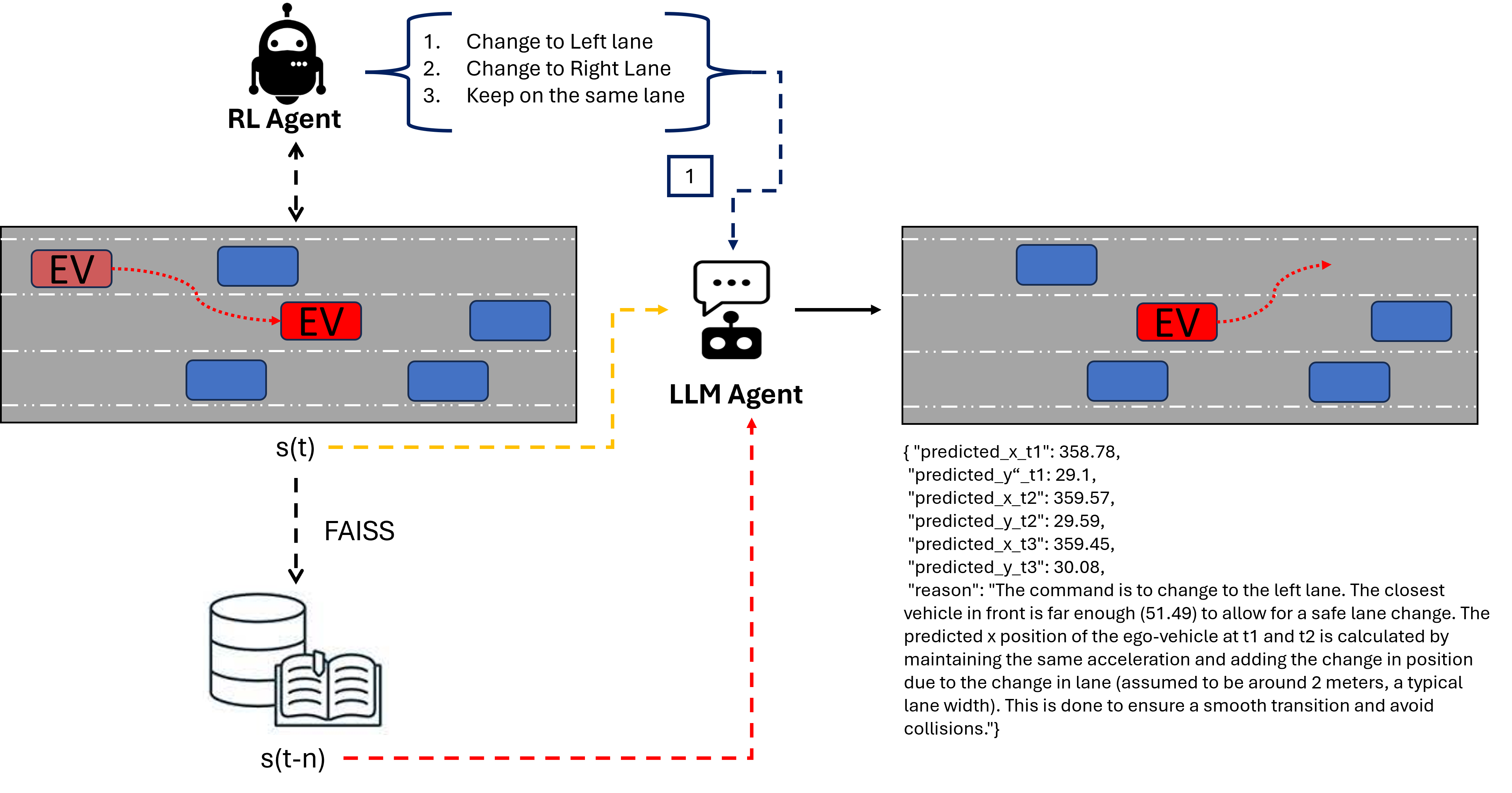}
\caption{LLM-based vehicle trajectory planning structure: The RL agent observes the traffic (surrounding vehicles) and provides a high-level action for a lane change. Then, the LLM agent retrieves the highD dataset by using FAISS and provides the next three trajectory points.}
\label{fig:RL_LLM}
\end{figure*}


\subsubsection{\textbf{Lateral Dynamics}}

The RL agent can perform three high-level actions: left lane change, right lane change, and staying in the current lane. These decisions are discrete. While the decision-making model decides to go left, right or stay in the lane, the PID controller calculates lateral acceleration as defined in equation \ref{PID}. It controls the ego vehicle to achieve a desired position.

\begin{equation}
u(t) = k_p e(t) + k_i \int_0^t e(\tau) d \tau + k_d \frac{d}{dt}e(t)
\label{PID}
\end{equation}

Where \(u(t)\) represents the acceleration at time t, \(k_p e(t)\) is the proportional term, \(k_i \int_0^t e(\tau) d\tau\) is the integral term, and \(k_d \frac{de(t)}{dt}\) is the derivative term. Here, \(e(t)\) is the error signal, indicating the difference between the desired position and its actual position at time \(t\). 

\subsubsection{\textbf{Longitudinal Dynamics}}
The Intelligent Driving Model (IDM) controls vehicle acceleration in a longitudinal direction, considering the distance and velocity of the preceding vehicle and the EV, and consequently alters the EV's velocity. The IDM aims to prevent collisions by observing distance and decreasing velocity if the distance is close to the threshold value. The specific parameters to calculate acceleration concerning the equation \ref{eq:IDM_a} is shown in Table \ref{tab:IDM} for the IDM model.

\begin{equation}
a=a_{\max }\left[1-\bigg(\frac{v}{v_{\text {desired }}}\bigg)^{4}-\left(\frac{s^{*}\left(v, v_{\text {lead }}\right)}{s}\right)^{2}\right]
\label{eq:IDM_a}
\end{equation}

\begin{table}[ht]\centering
\caption{ \\IDM Parameters}
\ra{1.3}
\begin{tabular}{@{}lccc@{}}\toprule
 
$Parameter$  & $Value$  \\ \midrule
Minimum Distance ($s_{0}$) & 10 m  \\
Desired Velocity ($V_{desired}$) & 130 km/h  \\
Maximum Acceleration ($a_{max}$) & 3  m/s  \\
Maximum Deceleration ($b_{max}$) & 5 m/s \\
Safe Deceleration ($b_{safe}$) & 4 m/s \\
\bottomrule
\end{tabular}
\label{tab:IDM}

\end{table}

\subsection{Language Model for Trajectory Planning} \label{trajectory_planning}
In our proposed approach, we use the reasoning and interpretability capabilities of an LLM to predict the future states which form the trajectory of the ego-vehicle. Firstly, we formulate the trajectory planning as a language problem where we define the inputs for our LLM agent as current state information denoted as \textit{S$_t$}. The \textit{S$_t$} includes current \( x \) and \( y \), along with the respective velocities \(v_{x}\) and \(v_{y}\), and accelerations \(a_{x}\) and \(a_{y}\), of the ego-vehicle. The second input to the LLM agent is the information about the surrounding vehicles at the current time step, which is denoted as \(S_{surrounding_t}\), which includes front and back sight distance with the preceding vehicle's longitudinal velocity. Lastly, we input the past similar trajectories for a given state, which is denoted as \(Hist_{t}\); see details in Sec \ref{RAG_agent}.

The trajectory prediction problem can be formulated as follows:

\begin{equation}
P_{States} = LLM(S_t, S_{surrounding_t}, Hist_{t})
\end{equation}
\noindent
where \(P_{States}\) represents the predicted state of the ego-vehicle at time \(t+1\), \(LLM\) denotes the LLM agent.

\noindent
In order to enforce the output of the LLM agent to be in a compatible format that can be converted into vehicle actions, we use the function calling tool of LLMs:

\begin{equation}
\begin{split}
pred\_x_{t+n} &= Func_x(S_t, S_{surrounding_t}, Hist_t), \\
pred\_y_{t+n} &= Func_y(S_t, S_{surrounding_t}, Hist_t), \\
\end{split}
\end{equation}
\noindent
where \( n \in \{1, 2, 3\} \). The pre-defined functions, \( Func_x \) and \( Func_y \), are designed to output the next lateral and longitudinal positions of the ego vehicle.
The output from the LLM agent is represented as follows:

\begin{equation}
P_{States} = \left\{
\begin{array}{l}
\textit{pred\_x}_{t+n}, \\
\textit{pred\_y}_{t+n}, \\
\textit{reason}
\end{array}
\right.
\end{equation}

The argument \(reason\) provides insights into the rationale behind the predicted states, offering explanations that enhance the system's interpretability. 

\begin{figure*}[ht]

\begin{mdframed}
\begin{center}
 \includegraphics[width=1\textwidth,height=\textheight,keepaspectratio]{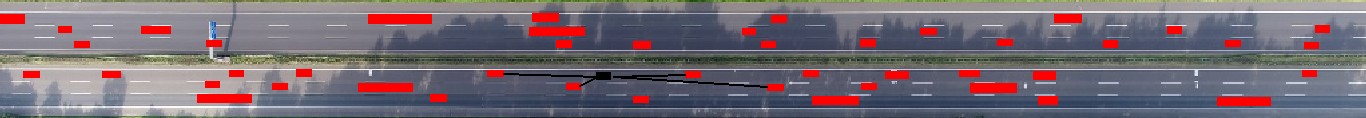}
\end{center}

\textbf{LLM role:} You are an AI assistant capable of predicting a vehicle's next position on a highway. You will be provided with commands to change to the left lane, right lane, or keep in the same lane. You are given information about the current velocity, acceleration, and position of the ego-vehicle. Additionally, you are provided with information about the front and back sight distance with other vehicles, including the preceding vehicle's x velocity. You are also given past similar trajectories for the given state information.

\textbf{LLM Prediction Functions:} Predict the next x,y coordinate of the ego-vehicle with respect to given 
Position Parameters (x, y, velocity, acceleration)

    \textbf{Prompt:} You are given the current state with x position \texttt{input\_params.current\_x}, y position \texttt{input\_params.current\_y}, x velocity \texttt{input\_params.velocity\_x}, y velocity \texttt{input\_params.velocity\_y}, acceleration in x \texttt{input\_params.acceleration\_x}, acceleration in y \texttt{input\_params.acceleration\_y}, front sight distance \texttt{input\_params.frontSightDistance}, back sight distance \texttt{input\_params.backSightDistance}, and preceding vehicle's x velocity \texttt{input\_params.precedingXVelocity}, along with similar past trajectories: \texttt{nearest\_neighbors\_str}. You are also given a command to \texttt{meta\_action}. Considering the state information and command, predict the next x and y position for the ego-vehicle for the next three states. You should make logical predictions to avoid collision with other vehicles, ensure safe travel, and smooth speed transitions. You should briefly explain your reason for the predictions made.
\end{mdframed}
\caption{Simulation platform and LLM prompt: The ego vehicle is shown as black, and other traffic vehicles from the dataset are depicted as red. Black lines represent the radar's detection of surrounding vehicles and measure the distance. LLM is given a role, and inputs are given with a parameters prompt, which returns an output as a predicted waypoint.}
\label{fig:LLM_prompt}
\end{figure*}

\subsection{Knowledge Retrieval for Trajectory Planning} \label{RAG_agent}
In our approach, we aimed to use the LLM as a zero-shot trajectory planner to evaluate its generalisability in autonomous driving task. However, our agent required some knowledge to provide reasonable predictions for future states. Therefore, we generated a large knowledge base made up of past trajectories of different vehicles in a highway. Our knowledge base is collected from the real vehicle trajectories of highD \cite{krajewski2018highd} dataset.
To provide a fast knowledge retrieval for our LLM agent, we adapted Facebook AI similarity search (FAISS)\cite{johnson2019billion} algorithm is a tool designed for efficient similarity search and clustering of dense vectors. Our knowledge retrieval has two main steps:

\begin{itemize}
    \item \textbf{Vectorisation} : We converted the historical trajectories stored within our knowledge base into vector representations. Similarly, for any given time step, the current state \textit{S$_t$} is also vectorized.
    \item \textbf{Retrieval} : We use L2 distance metric, to retrieve the most relevant three historic trajectories that closely match with the given \textit{S$_t$} 
\end{itemize}

\subsection{Prompt Engineering}
In order to get a desired output from our LLM agent, we used prompt engineering technique with the function calling to restrict the output of the system. We instructed the LLM to consider the given state information along with the past similar trajectory and asked it to predict the next states by \textit{avoiding any collisions} \textit{ensuring safe travel} and \textit{smooth speed transitions}. The role and prompt messages given to our LLM agent is shown in Figure \ref{fig:LLM_prompt}.

\begin{figure*}[h]
\centering
\includegraphics[width=0.9\textwidth,height=\textheight,keepaspectratio]{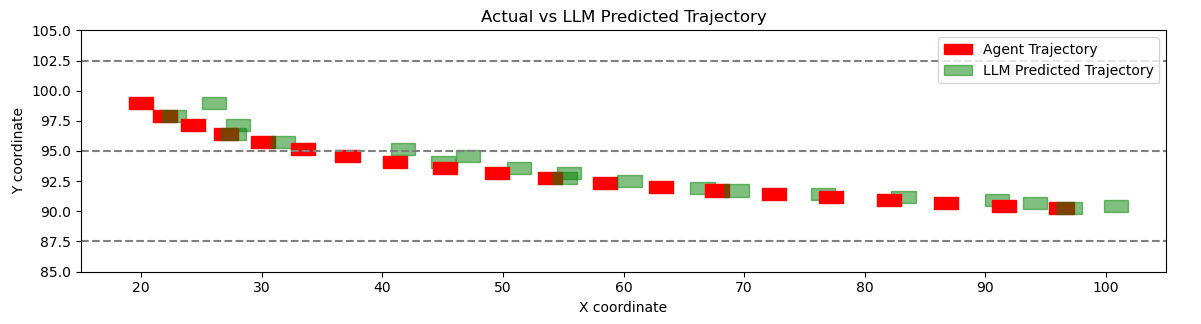}
\caption{RL initiated a lane change, LLM provided trajectory waypoints for lane change. Red vehicles shows actual vehicle trajectory, green vehicle shows LLM predicted waypoint for next three step.}
\label{fig:LLM_trajectory}
\end{figure*}

\section{Experiments}

We used the highD traffic dataset \cite{krajewski2018highd} to evaluate our proposed approach. We simulated our experiments using Pygame as shown in Figure \ref{fig:LLM_prompt}, where the ego-vehicle is shown as a black rectangle and the other vehicles in the traffic as a red rectangles. In our experiments, we aim to evaluate the performance of LLM for driving safety and decision-making. Firstly, we run the LLM as a trajectory planner and measure its number of collisions, average velocity and inference time. Secondly, we integrate the LLM agent as a safety layer alongside the RL agent. We assess LLM's ability to improve decision-making through the reasoning process in evaluating and potentially opposing or supporting the RL agent's proposed actions.

\subsection{Trajectory Planning}

In this experiment, the LLM agent is tasked with predicting the next trajectory points of the ego vehicle, guided by high-level action input from the RL agent. The LLM agent is also provided with the current state information and retrieves the three most relevant historical trajectories from a database. This prediction process becomes particularly crucial when the RL agent gives a lane change signal. The 'keep on the same lane' action does not necessitate explicit trajectory planning—the primary focus is on the accurate prediction of lane change trajectory points. We compare the actual trajectory with the LLM-predicted trajectory to assess the accuracy of LLM's predictions. 



\begin{table*}[ht]\centering
\caption{\\ Comparison of $RL+IDM$, $RL+LLM_{Trajectory}$ and $RL+IDM+LLM_{safety}$}
\ra{1.3}
\begin{tabular}{@{}rcccccccc@{}}\toprule
& \multicolumn{1}{c}{$RL+IDM$} & \phantom{abc}& \multicolumn{2}{c}{$RL+LLM_{Trajectory}$} &
\phantom{abc} & \multicolumn{3}{c}{$RL+IDM+LLM_{Safety}$}\\
\cmidrule{2-2} \cmidrule{4-5} \cmidrule{7-9}
Collision No &3.84   & &  &2   &&&0.28     &  \\
Velocity (km/h) &100.8   & &  &121.28   &&&111.34     &  \\
Inference Time (s) & 0.002 & &&6.79   &  &&2.89& \\    

\bottomrule
\end{tabular}
\label{compare_methods}
\end{table*}

\subsection{Safety Check}

In the second experiment, we explore the LLM agent as a safety layer for RL's policy before executing any actions. The LLM predicts the high-level actions alongside and without having direct input from the RL agent. The LLM uses its reasoning capability and the given set of instructions to make safe decisions at the given current state. The experiment is designed on a consensus-based action execution model, where the action is only executed when both the LLM and RL agents independently decide on the same action. Otherwise, the system defaults to maintaining the current lane and avoiding potential manoeuvre risks such as lane change or overtaking.

\section{Results \& Discussion}

We tested our proposed method on highway driving scenarios extracted from an actual highway driving dataset and compared it with an RL approach as a baseline. Table \ref{compare_methods} shows the results for the mentioned approaches. 

Considering the role of the LLM agent as a trajectory planner, our proposed approach performed an average 2.0 number of collisions with an average velocity of 121.23 km/h. In contrast, the baseline resulted in an average of 3.84 collisions with an average velocity of 100.8 km/h. When the LLM agent acted as a safety layer, the number of collisions that occurred was an average of 0.28, with an average velocity of 111.34 km/h. 

Considering HighwayLLM as a trajectory planner, the LLM generates inaccurate trajectory points at the initial states due to a lack of information and understanding. However, as the lane change initiates, the model provides smoother waypoints. Figure \ref{fig:LLM_trajectory} illustrates a visual comparison of LLM and actual trajectories. In this figure, the exact position of the ego vehicle is represented by the red vehicle, whereas the green vehicle indicates the trajectory points provided by the LLM. 

We showed LLMs reasoning in Figure \ref{fig:graybox_reasoning} for a lane change action in 6 frames. This figure represents LLM's reasoning capability while RL is providing high-level action; LLM provides a trajectory waypoint and explains the scenario and action based on its understanding. LLM's reasoning could be used to explain RL's black-box nature for autonomous driving.



We measured the LLM agent's response time for safety and trajectory point generation. It is 6.79 seconds on average for the $LLM_{trajectory}$, while the $LLM_{safety}$'s action generation took only 2.89 seconds.This response time depends on the size of retrieval database and the number of output trajectory points required, such as three pairs of lateral and longitudinal (x, y) coordinates. On the other hand, $LLM_{safety}$ provides only one output as a lane change action, while trajectory points output in total six numbers of data as x,y couples; therefore, it responds more than two times faster than generating trajectory points. Comparing LLM's response time, RL provides an action with an average of 2 m seconds.

Although these results highlight the benefits of integrating language models with RL methods for highway driving in terms of vehicle velocity and safety improvement, LLM's effectiveness needs to be tested in various tracks and datasets. In addition, we did not test LLM by itself for highway driving due to its well-known hallucination issues. Therefore, we either used LLM by giving an RL command or as a safety layer in which LLM and RL agreed on the same decision.

\section{Conclusion}

This paper proposes an RL-informed LLM for highway driving and compares its performance with the state-of-the-art RL method. Our proposed HighwayLLM agent is employed as a safety layer in highway driving to evaluate the decisions made by the pre-trained RL agent, and it acts as a trajectory planner by cooperatively working with the RL to make safer decisions in complex driving environments. We compared the safety, velocity, and response time of HighwayLLM with a baseline RL approach, and our experiments showed that HighwayLLM provides safer driving and results in fewer collisions, albeit with slightly an increased response time due to its understanding of the current state information, reasoning and predicting the future waypoints for the ego-vehicle. As a next step, we aim to develop a trainable end-to-end pipeline to fine-tune the LLM with RL feedback for future work.


%

\begin{tcolorbox}[fonttitle=\bfseries, title=]
\begin{center}
  \includegraphics[width=\textwidth,height=\textheight,keepaspectratio]{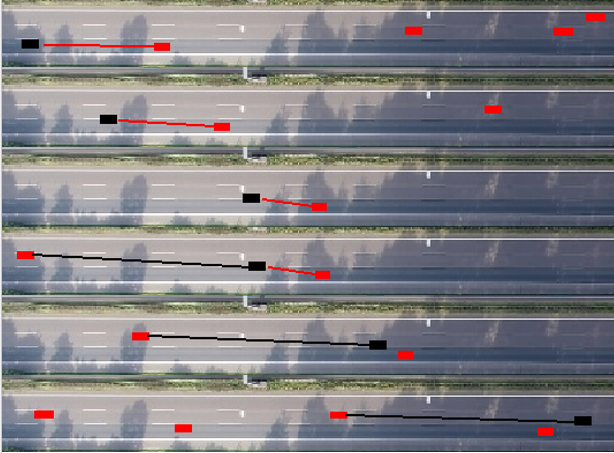}

\label{fig:reasoning}
\end{center}

    \textbf{\texttt{1st}} The command is to change to the left lane. The current y position is 97.09, so in order to safely change lanes and remain between the highway boundaries of 72 and 104, I decreased the y position slightly in each of the next 3 states to ensure that even if the preceding vehicle changes lanes or adjusts their speed, our ego-vehicle will still have enough room for safe maneuvering and speed transitions.\\
    \textbf{\texttt{2nd}} To safely change lanes to the left, the ego-vehicle should decrease its y-position. Since the command is to change to the left lane, and considering the highway boundaries for the y-position (72 and 104), adjusting the y-position to 95.5 seems a reasonable choice. It ensures safe travel, avoids collisions with other vehicles, and enables smooth speed transitions.\\
    \textbf{\texttt{3rd}} The command is to change to the left lane, and the given front sight distance is 48.06. Therefore, we predict that the ego-vehicle will move to the left lane and decrease its y-position to maintain a safe distance from the preceding vehicle while staying within the highway boundaries (72 and 104).\\
    \textbf{\texttt{4th}} The preceding vehicle is moving slower than the ego-vehicle, so there is no immediate danger of collision. The command is to change to the left lane, so the y-position of the ego-vehicle should be adjusted to a lower value while ensuring it remains within the highway boundaries (72 and 104). The x-position should smoothly increase to maintain safe travel and ensure smooth speed transitions.\\
    \textbf{\texttt{5th}} I have decreased the predicted y-position for each of the next 3 states ($t1_{y}$ and $t2_{y}$) to safely move to the left lane, while staying within the highway boundaries (72 to 104) for y-position.\\
    \textbf{\texttt{6th}} The current velocity and acceleration of the ego-vehicle suggest that it will continue to move in the same direction. 
\end{tcolorbox}
\captionof{figure}{RL initiated a lane change, and LLM provides trajectory waypoints for lane change and reasoning for lane change action. Black vehicle shows ego vehicle, and red vehicles are surrounding vehicles.}\label{fig:graybox_reasoning}

\bibliographystyle{unsrt}
\bibliography{library}
\end{document}